\documentclass[10pt,twocolumn,letterpaper]{article}

\usepackage{cvpr}              

\usepackage{graphicx}
\usepackage{amsmath}
\usepackage{amssymb}
\usepackage{booktabs}
\usepackage[pagebackref,breaklinks,colorlinks]{hyperref}

\usepackage[capitalize]{cleveref}
\crefname{section}{Sec.}{Secs.}
\Crefname{section}{Section}{Sections}
\Crefname{table}{Table}{Tables}
\crefname{table}{Tab.}{Tabs.}

\def\cvprPaperID{*****} 
\def\confName{CVPR}
\def\confYear{2023}

\usepackage{xspace}
\usepackage{multirow}
\usepackage{booktabs}
\usepackage[accsupp]{axessibility}  

\DeclareMathOperator*{\argmax}{arg\,max}
\DeclareMathOperator*{\argmin}{arg\,min}

\newcommand{\featvec}[3]{\ensuremath{#1^{(#2,#3)}}}

\begin{document}

\title{Sanity checks for patch visualisation in prototype-based image classification}

\author{Romain Xu-Darme$^{1,2}$, Georges Quénot$^2$, Zakaria Chihani$^1$, Marie-Christine Rousset$^2$\\ \\
$^1$ Université Paris-Saclay, CEA, List, F-91120, Palaiseau, France\\
$^2$ Univ. Grenoble Alpes, CNRS, Grenoble INP, LIG, F-38000 Grenoble, France \\
{\tt\small \{romain.xu-darme, zakaria.chihani(at)cea.fr\}}\\
{\tt\small \{georges.quenot, marie-christine.rousset(at)imag.fr\}}\\
}
\maketitle
\begin{abstract}
	In this work, we perform an analysis of the visualisation methods implemented in ProtoPNet and ProtoTree, two self-explaining visual classifiers based on prototypes. 
We show that such methods do not correctly identify the regions of interest inside of the images, and therefore do not reflect the model behaviour, which can create a false sense of bias in the model.
We also demonstrate quantitatively that this issue can be mitigated by using other saliency methods 
that provide more faithful image patches. 
\end{abstract}
\section{Introduction}\label{sec:intro}
\begin{figure}[htbp]
 \caption{\textbf{Explanations of a ProtoTree} when using different visualisation methods. Due to the imprecision of upsampling when visualising both the prototype (right) and the part in the test image (left), the user might deduce that the model is comparing tree branches, when it is actually also taking the bird into account.}
\begin{subfigure}[hb]{0.30\linewidth}
	\centering
	\includegraphics[width=\textwidth]{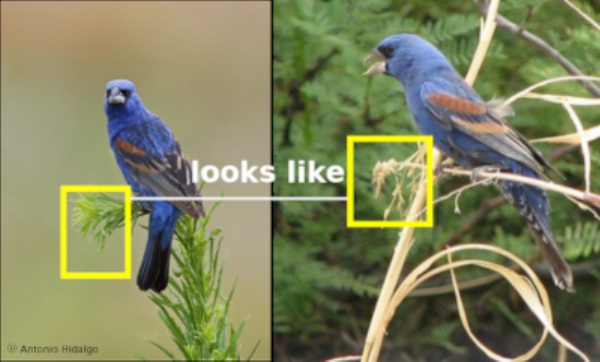}
	\caption{Upsampling.}\label{fig:inference_upsampling}
\end{subfigure}
\hfill
\begin{subfigure}[hb]{0.30\linewidth}
	\centering
	\includegraphics[width=\textwidth]{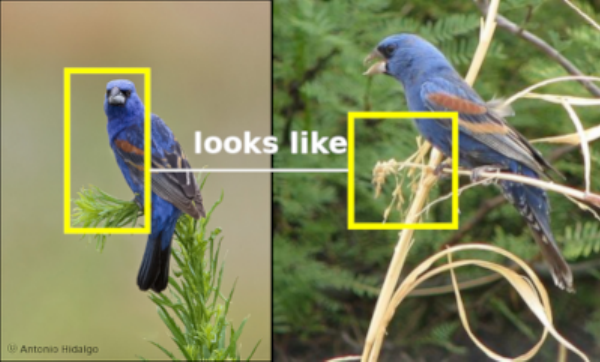}
	\caption{SmoothGrads.}
\end{subfigure}
\hfill
\begin{subfigure}[hb]{0.30\linewidth}
	\centering
	\includegraphics[width=\textwidth]{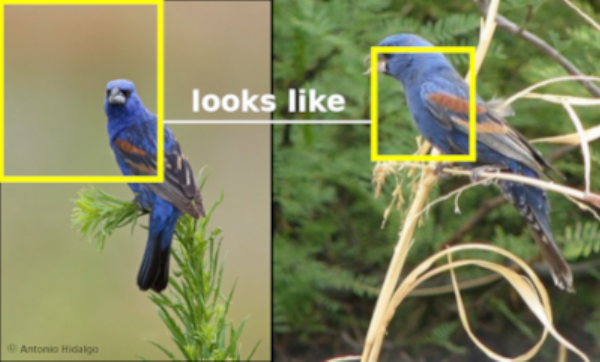}
	\caption{PRP}
\end{subfigure}
  \label{fig:comparisons}
  \vspace{-0.7cm}
\end{figure}

During the last decade, the field of Explainable AI (XAI)
has gained wide-spread recognition among the scientific community~\cite{bodria2021benchmarking,nauta2023anecdotal}.
One major avenue of research in this field consists in developing architectures and training procedures such that the resulting model should be \textit{self-explaining}.
In computer vision, such architectures often use a case-based reasoning mechanism~\cite{chen2019this,nauta2021neural,rymarczyk2021protopshare,rymarczyk2021interpretable,donnelly2021deformable} where new instances of a problem are solved and explained using comparisons with visual examples (\textit{prototypes}) extracted from the training data.
In particular, ProtoPNet~\cite{chen2019this} and ProtoTree~\cite{nauta2021neural} have shown that self-explaining architectures can reach accuracy levels comparable to more opaque models on fine-grained recognition tasks~\cite{welinder2010caltech,krause20133d}.
However, both models sometimes produce explanations using image patches that seem to be focused on elements unrelated to the object itself (Fig.\ref{fig:inference_upsampling}).
Importantly, 
a prototype focusing on the background might indicate a 
\textit{systemic} bias in the model and seriously hinder the user's trust in it.
However, recent work~\cite{gautam2022this} indicates that there exists in reality an imprecision in the patch visualisation method implemented in ProtoPNet. More generally, imprecise visualisation methods may suggest model bias where there is none, while sometimes hiding more systemic issues.
\paragraph{Our contribution:} In this work, we perform an analysis of the visualisation methods implemented in ProtoPNet and ProtoTree, answering the following research questions: do these architectures generate faithful explanations reflecting their decision-making process? do they produce decisions based on relevant parts of the image? We confirm quantitatively the results of \cite{gautam2022this} on ProtoPNet and show that ProtoTree also generates imprecise visual patches. Additionally, using the object segmentation provided in the CUB-200-2011 dataset, we propose a new relevance metric and show that in both architectures, such imprecise visualisations often create a false sense of bias that is largely mitigated by the use of more faithful methods. Finally, we discuss the implications of our findings to other prototype-based models sharing the same visualisation method.

\section{Related work}\label{sec:related_work}
\paragraph{Prototype-based classifiers}
first encode images into a high dimensional feature space (\textit{latent space}) - generally using the first layers of a pre-trained convolutional neural network (CNN) as the backbone of the encoder.
During training, these classifiers extract a set of reference feature vectors and their visual counterparts from the training set, called \textit{part prototypes} (for simplicity, we simply use the term \textit{prototypes}). Prototypes are either discriminative of a particular class~\cite{chen2019this,donnelly2021deformable,wang2021interpretable}
or shared among multiple classes~\cite{nauta2021neural,rymarczyk2021interpretable,rymarczyk2021protopshare}.
During inference, the similarity between a given prototype and the test image is computed using the L2-distance (or cosine distance~\cite{wang2021interpretable}) between their respective latent representations. 
In the case of ProtoPNet~\cite{chen2019this}, ProtoPShare~\cite{rymarczyk2021protopshare}, ProtoPool~\cite{rymarczyk2021interpretable}, Deformable ProtoPNet~\cite{donnelly2021deformable} and TesNet~\cite{wang2021interpretable},
all similarity scores are then processed through a fully connected layer to produce the prediction. In the case of ProtoTree~\cite{nauta2021neural}, they are used to compute a path across a soft decision tree where each leaf corresponds to a categorical distribution among classes. \\
Note that although our study focuses on ProtoPNet and ProtoTree, all the aforementioned methods (ProtoPShare, ProtoPool, Deformable ProtoPNet, TesNet) share a common code base inherited from ProtoPNet and therefore are theoretically susceptible to the issue raised in this contribution. \\
\textbf{Saliency methods}
aim at identifying the most important pixels of an image \wrt the output of a given neuron.
Gradient-based approaches~\cite{simonyan2013deep} compute the partial derivative of the target neuron output \wrt to each input pixel, with improvements such as Integrated Gradients~\cite{sundararajan2017axiomatic}, gradients $\odot$ input~\cite{shrikumar2017learning} (with $\odot$ denoting the element-wise multiplication), or SmoothGrads~\cite{smilkov2017smoothgrads}.
In particular, SmoothGrads ``adds noise to remove noise'' by averaging gradients over noisy copies of the input image.
For ProtoPNet, \cite{gautam2022this} proposes a variant of LRP~\cite{binder2016layer} called Prototype Relevance Propagation (PRP), implementing a dedicated rule to propagate relevance across the layer in charge of computing similarity scores.
Since
Integrated Gradients, LRP$-\epsilon$ and Deep-LIFT~\cite{shrikumar2017learning} are all equivalent to gradient~$\odot$~input for CNNs based on ReLU activation~\cite{ancona2017unified,shrikumar2016not},
in this work we choose to compare the original part visualisation generated by ProtoPNet and ProtoTree to visualisations generated using SmoothGrads $\odot$ input and PRP.
Note that we exclude Guided-Backpropagation~\cite{springenberg2014striving} and its application to GRADCAM~\cite{selvaraju2016grad} due to the results of \cite{adebayo2018sanity} which raise some concerns regarding the faithfulness of the saliency maps generated by these methods.\\
\textbf{Evaluation metrics:}
We first focus on the property known as \textit{faithfulness}~\cite{alvarezmelis2018towards}
which quantifies the adequacy between a saliency method and the model behaviour. Faithfulness can be evaluated by model parameter randomisation~\cite{adebayo2018sanity} or deletion/insertion methods, which monitor the evolution of a neuron's output when the most/least important pixels of the input image are removed incrementally~\cite{petsiuk2018rise,tomsett2019sanity} or individually~\cite{alvarezmelis2018towards}. Deletion/insertion metrics check the ability of a saliency method to correctly \textit{sort} pixels by importance \wrt to a given neuron's output, \ie
``removing'' the most salient pixels identified by a faithful saliency method should result in a strong variation of the neuron's output. \\
Secondly, we wish to evaluate the \textit{relevance} of prototype-based explanations. \cite{nauta2020this} applies perturbations (\eg changes in colour or shape) on images and monitors the evolution of the similarity score for each prototype. Note that since these perturbations are applied on the \textit{entire} image, this method does not require a precise location of the image patches.

\section{Theoretical background}\label{sec:theory}
In this work, we consider a classification problem with a training set $X_{train} \subseteq \mathcal{X}\times \mathcal{Y}$. Let $f$ be a fully convolutional neural network (fCNN) processing images in $\mathcal{X}$ and producing a $D$-dimensional latent representation of size $(H,W)$. For $x\in \mathcal{X}$, we denote $\featvec{f}{h}{w}(x)\in \mathbb{R}^D$ the
vector corresponding to the $h$-th row and $w$-th column of $f(x)$.

\subsection{ProtoPNet and ProtoTree}
\begin{figure}[htbp]
    \centering
    \caption{\textbf{From a similarity map to a part visualisation}. A saliency method generates a saliency map from the similarity map. Then, we retain only the most salient pixels through thresholding and crop the original image to produce a part visualisation.}
    \label{fig:part_visualisation}
    \includegraphics[width=\linewidth]{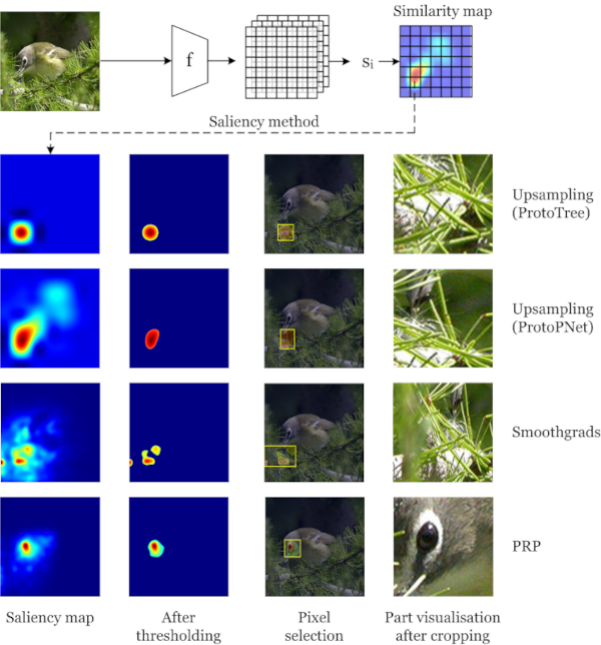}
\end{figure}

\paragraph{Learning prototypes}
For $x\in \mathcal{X}$, ProtoPNet and ProtoTree compute their decision (classification) based on similarities between the latent representation $f(x)$ and a set of feature vectors $(r_1, \ldots r_p)$ that are learned during training and act as reference points in the latent space. More precisely, the similarity between $r_i$ and a particular vector $\featvec{f}{h}{w}(x)$ is defined as
$\featvec{s_i}{h}{w}(x)= log\big((\|\featvec{f}{h}{w}(x)-r_i\|_2^2+1)/(\|\featvec{f}{h}{w}(x)-r_i\|_2^2+\epsilon)\big)$ (ProtoPNet) or $\featvec{s_i}{h}{w}(x)=e^{-\|\featvec{f}{h}{w}(x)-r_i\|_2^2}$ (ProtoTree), where $\|.\|_2$ denotes the L2 distance.
$s_i(x)\in \mathbb{R}^{H \times W}$ is called the \textit{similarity map} between $x$ and $r_i$.
The model decision process $d$ is a function of the aggregation of high similarity scores $s(x)=(\max(s_1(x)), \ldots \max(s_p(x)))$: weighted sum for ProtoPNet, soft decision tree for ProtoTree. During training, the parameters of the feature extractor $f$, of the decision function\footnote{The details of the decision function $d$ are not relevant to our work, which focuses on the method used by ProtoPNet and ProtoTree to build a saliency map out of the similarity map $s_i(x)$.} $d$, and the reference points $r_i$ are jointly learned in order to minimize the cross-entropy loss between the prediction $d \circ s(x)$ and the label $y$, for $(x,y)\in X_{train}$.
After training, the reference points $r_i$ are ``pushed'' toward latent representations of parts of training images, in a process called \textit{prototype projection}. More formally, a prototype $P_i$ is a tuple $P_i=(p_i, h_i, w_i, r_i)$, computed using the training set $X_{train}$, where
\begin{equation}
	\left\{
	\begin{array}{l}
		p_i, h_i, w_i = \argmin\limits_{x\in X_{train}, h, w} \|\featvec{f}{h}{w}(x)-r_i\|_2^2 \\
		r_i \leftarrow \featvec{f}{h_i}{w_i}(p_i)
	\end{array}
	\right.
\end{equation}
Thus, prototype projection moves each $r_i$ to a nearby point that is, by construction, obtained from an image $p_i$. 
%
\paragraph{From similarity map to part visualisation.}
Given an image $x$ and a prototype $P_i$, ProtoPNet generates a visualisation of the most similar image patch in $x$ by upsampling the similarity map $s_i(x)\in \mathbb{R}^{H\times W}$ to the size of $x$ using cubic interpolation, then cropping the resulting saliency map to the 95th percentile. ProtoTree retains only the location of highest similarity in $s_i(x)$ before upsampling (setting all other locations to $0$), as shown in Fig.~\ref{fig:part_visualisation}. Note that the same method is also used to visualise the prototype itself by setting $x=p_i$.
However, neither approach factors in the size of the receptive field\footnote{It is actually computed in the code of ProtoPNet, but never put to use.} of each neuron in the last layers of a CNN. Indeed, 
the value $\featvec{s_i}{h}{w}(x)$ may actually depend on the entire image $x$ rather than a localized region~\cite{araujo2019computing}. \\
Finally, during inference, for $x\in \mathcal{X}$ and for each prototype $P_i$, both ProtoPNet and ProtoTree find the vector in $f(x)$ closest to $r_i$, corresponding to the highest score of the similarity map $s_i(x)$. If this score is above a given threshold, then they show side by side patches of images extracted from the prototype image $p_i$ and $x$ (\textit{this looks like that}).

\subsection{Improving part visualisation}
In this paper, we wish to showcase the benefits of using other saliency methods for generating part visualisation.
Similar to ProtoTree, for $x\in \mathcal{X}$ and a prototype $P_i$, we first find the coordinates of the highest similarity score
\begin{equation}
    \left\{
	\begin{array}{l}
        \mathbf{H}_i(x),\mathbf{W}_i(x) = \argmax\limits_{h,w} \featvec{s_i}{h}{w}(x) \\
        \mathbf{S}_i(x) = \max\limits_{h,w} \featvec{s_i}{h}{w}(x)=\featvec{s_i}{\mathbf{H}_i(x)}{\mathbf{W}_i(x)}(x)
	\end{array}
	\right.
\end{equation}
We generate saliency maps
by applying  SmoothGrads $\odot$ input or PRP on the output of the neuron $\featvec{s_i}{\mathbf{H}_i(x)}{\mathbf{W}_i(x)}$. Then, we obtain a part visualisation by retaining only the 2\% of most important pixels from $x$ and cropping the image accordingly (Fig.~\ref{fig:part_visualisation}). Again, the same method is applied in order to extract a part visualisation for each prototype.

\subsection{Measuring faithfulness}\label{sec:faithfulness}
Similar to~\cite{petsiuk2018rise}, $\forall x\in \mathcal{X}$ and $\forall P_i$, we compare the faithfulness of the saliency methods by
computing the Area Under the Deletion Curve (AUDC)
\begin{equation}
\int\limits_0^{a_{max}}\tau(a)da=\int\limits_0^{a_{max}}\dfrac{\mathbf{S}_i(x)}{\featvec{s_i}{\mathbf{H}_i(x)}{\mathbf{W}_i(x)}(x \odot m_a(x))}da
\end{equation}
where  $m_a(x)$ is the binary mask obtained after ``deleting'' (colouring in black) the $a$\%
most salient pixels in $x$. $\tau(a)$ measures the relative drop in the similarity scores between the original image $x$ and a perturbed input $x \odot m_a(x)$ at the location $(\mathbf{H}_i(x),\mathbf{W}_i(x))$ inside the similarity map $s_i$.
$\tau(a) \approx 1$ indicates that these pixels have no impact on the similarity score (unfaithful saliency method), while $\tau(a) \approx 0$ indicates that these pixels have
a high impact on the similarity score, thus that the saliency method correctly identifies relevant pixels \wrt to the similarity score (faithful method). In order to reduce unexpected behaviours from the CNNs~\cite{gomez2022metrics}, we restrict ourselves to deleting a small portion of the original image ($a_{max}=$2\%).
Note that $\tau(a) < 0.2$ indicates that the deleted pixels amount to 80\% of the similarity score, thus that the area of \textit{the effective receptive field} of $f$ \wrt to $\mathbf{S}_i(x)$ is probably close to $a$.

\section{Experiments}\label{sec:experiments}
\begin{table*}[htbp]
\centering
\caption{\textbf{Average AUDC} of prototypes (left value) and test patches (right value) generated by ProtoPNet and Prototree when using upsampling, SmoothGrads, PRP and RandGrads. For each architecture/dataset, values in \textbf{bold} indicates the most faithful saliency method.
}\label{tab:faithfulness}
\centering
	{\renewcommand\baselinestretch{1.3}\selectfont \resizebox{\linewidth}{!}{
\vspace{0.3cm}
	\begin{tabular}{@{\hskip 8pt}c@{\hskip 8pt}|@{\hskip 8pt}c@{\hskip 8pt}|@{\hskip 8pt}c@{\hskip 8pt}|@{\hskip 8pt}c@{\hskip 8pt}c@{\hskip 8pt}c@{\hskip 8pt}c@{\hskip 8pt}}
		\toprule
		\multirow{2}{*}{Model} & \multirow{2}{*}{Backbone} & \multirow{2}{*}{Dataset} & \multicolumn{4}{c}{Method} \\
		& & & Upsampling & PRP & SmoothGrads & RandGrads\\
		\midrule
		\multirow{3}{*}{ProtoPNet} & VGG19    & CUB-c & 0.41 ($\pm$ 0.12) / 0.74 ($\pm$ 0.19) & 0.39 ($\pm$ 0.10) / 0.70 ($\pm$ 0.19) & \textbf{0.37 ($\pm$ 0.11) / 0.68 ($\pm$ 0.20)} &  0.77 ($\pm$ 0.18) /  0.94 ($\pm$ 0.12)\\
		\cline{2-7}
                                   & \multirow{2}{*}{ResNet50} & CUB-c & 0.39 ($\pm$ 0.18) / 0.71 ($\pm$ 0.28) & \textbf{0.31 ($\pm$ 0.14) / 0.61 ($\pm$ 0.28)} & 0.37 ($\pm$ 0.18) / 0.66 ($\pm$ 0.29) & 0.77 ($\pm$ 0.23) / 0.93 ($\pm$ 0.18) \\
                                                  	         & & CARS & 0.46 ($\pm$ 0.19) / 0.88 ($\pm$ 0.20) & \textbf{0.31 ($\pm$ 0.11) / 0.68 ($\pm$ 0.24)} & 0.34 ($\pm$ 0.13) / 0.71 ($\pm$ 0.24) &  0.60 ($\pm$ 0.18) / 0.94 ($\pm$ 0.14)\\
        \midrule
        \multirow{2}{*}{ProtoTree} & \multirow{2}{*}{ResNet50} & CUB & 0.98 ($\pm$ 0.06) / 0.95 ($\pm$ 0.16) & \textbf{0.78 ($\pm$ 0.28) / 0.65 ($\pm$ 0.34)} & 0.91 ($\pm$ 0.22) / 0.82 ($\pm$ 0.28) & 0.99 ($\pm$ 0.08) / 0.98 ($\pm$ 0.10) \\
                                                  	         & & CARS & 0.96 ($\pm$ 0.13) / 0.91 ($\pm$ 0.21) & \textbf{0.75 ($\pm$ 0.25) / 0.71 ($\pm$ 0.30)} & 0.88 ($\pm$ 0.21) / 0.84 ($\pm$ 0.25) & 0.99 ($\pm$ 0.05) / 0.97 ($\pm$ 0.12) \\
        \bottomrule
	\end{tabular}
	}}
\end{table*}
We perform our experiments on two popular datasets for fine-grained recognition:
CUB-200-2011~\cite{welinder2010caltech} (CUB)
and Stanford Cars~\cite{krause20133d} (CARS).
For ProtoPNet, we use the images of the CUB dataset cropped to the object bounding box (we denote this dataset CUB-c) during training and inference.
We primarily use a Resnet50~\cite{he2016residual} backbone, pretrained on the iNaturalist~\cite{horn2017inaturalist} dataset (CUB) or the ImageNet~\cite{deng2009imagenet} dataset (CARS), with images of size $224\times 224$. To compare results with a different backbone, we also train a ProtoPNet on CUB-c using a VGG19~\cite{simonyan2015very} network. \\
For PRP, we use the code kindly provided by the authors. For SmoothGrads, we use 10 noisy samples per image and a noise level of $0.2$. In order to provide a clear baseline for the evaluation of the faithfulness and relevance of saliency maps, we also implemented a trivial method RandGrads returning a random saliency map drawn from a uniform distribution. For all three methods, we post-process the saliency map by averaging the gradients at each pixel location across the RGB channels, taking the absolute value (putting equal emphasis on positive and negative gradients), and applying a $5\times 5$ Gaussian filter in order to avoid isolated gradients due to pooling layers inside of $f$.
\paragraph{Faithfulness of patch visualisation.}
We measure the AUDC when visualising both prototypes and patches of images from the test set during inference:
for ProtoTree, we apply the saliency method only when the prototype is considered ``present'' (right branch of each decision node); for ProtoPNet, we apply the saliency method on the 10 patches of the test image most similar to any prototype of the inferred class. The AUDC score is approximated by averaging the similarity ratio $\tau(a)$ for deletion areas between 0\% and 2\%, with an increment value of 0.1\% (Table~\ref{tab:faithfulness}).
In all cases, the upsampling method used in ProtoPNet and ProtoTree leads to a higher AUDC score than SmoothGrads or PRP. In the particular case of ProtoTree, it is only marginally better than our random baseline RandGrads. This confirms the imprecision pointed out in \cite{gautam2022this} and \textit{extends the issue to ProtoTree}. Moreover, in general PRP seems to provide a more faithful saliency maps than SmoothGrads (lower AUDC).
Note that AUDC scores are fairly similar across methods (except RandGrads) on the CUB-c dataset, which is probably due to the image cropping that increases the relative area of the bird inside of the image and decreases the probability to miss the important pixels.
\begin{figure}[htbp]
	\caption{\textbf{Average similarity ratio v. deletion area} when using PRP for visualising prototypes. Best viewed in colour.}
    \centering
    \includegraphics[width=0.9\linewidth]{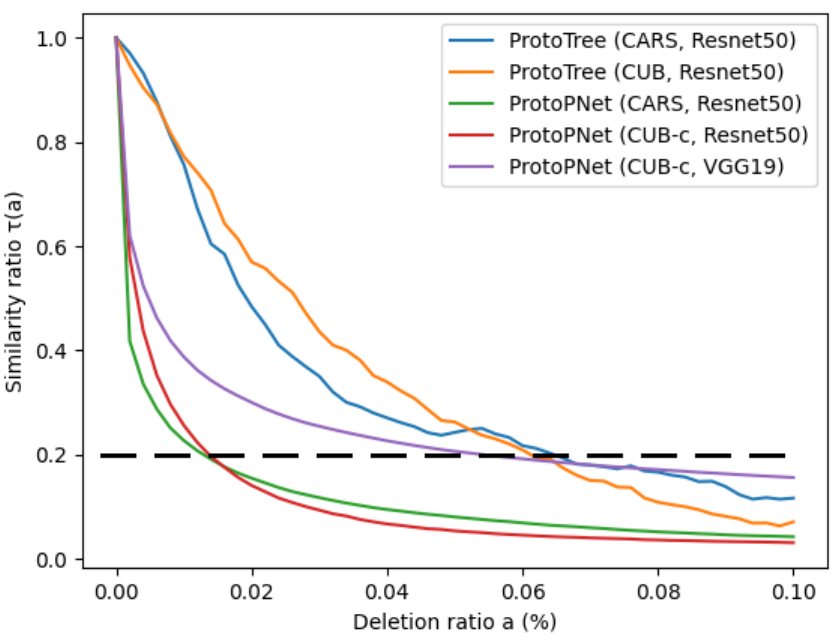}
    \label{fig:deletion_curve_prp}
    \vspace{-0.2cm}
\end{figure}
When extending the deletion area to 10\% of the image, we note that on average, the drop in similarity ratio ($\tau(a) < 0.2$) occurs below 2\% for ProtoPNet prototypes when using ResNet50, and around 7\% for ProtoTree prototypes or ProtoPNet using VGG19 (Fig.~\ref{fig:deletion_curve_prp}). This indicates that the size of the effective receptive field depends not only on the choice of backbone (VGG, ResNet50), but also on the model (ProtoTree, ProtoPNet).
For ProtoTree, this may be due to the fact that the decision tree shares prototypes among all classes and therefore does not focus on very small details, contrary to ProtoPNet\footnote{This effect is also present when using SmoothGrads and seems uncorrelated to the depth of the prototype inside the tree}.
Moreover, this clarifies the discrepancy in AUDC scores between ProtoPNet and ProtoTree visualisations. \\
\textbf{Relevance of patch visualisation.}
To evaluate the relevance of image patches (prototypes or test patches),
we measure their intersection with the object segmentation, assuming that such information is available (CUB dataset). If less than 5\% of the image patch
intersects the object segmentation,
then we consider it irrelevant, as it mostly focuses on the background.
\begin{table}[htbp]
\centering
\caption{\textbf{Measuring relevance}. Percentage of prototypes (left value) and test patches (right value) with less than 5\% of intersection with the object on the CUB dataset.}
	\label{tab:relevance}
	\centering
	{\renewcommand\baselinestretch{1.3}\selectfont \resizebox{\linewidth}{!}{
	\vspace{0.3cm}
	\begin{tabular}{@{\hskip 8pt}c@{\hskip 8pt}|@{\hskip 8pt}c@{\hskip 8pt}|@{\hskip 8pt}cccc}
		\toprule
		\multirow{2}{*}{Model} & \multirow{2}{*}{Backbone} & \multicolumn{3}{c}{Method} \\
		& & Upsampling & PRP & SmoothGrads \\
		\midrule
		\multirow{2}{*}{ProtoPNet} & VGG19    & 15.4\% / 23.3\% & 10.0\% / 16.6\% & 11.3\% / 19.9\% \\
                                   & ResNet50 & 2.1\% / 8.8\% & 1.4\% / 8.0\% & 0.9\% / 6.1\% \\
        \midrule
        ProtoTree & ResNet50 &  35.4\% / 51.9\% & 0.5\% / 0.5\% & 8.7\% / 14.5\% \\
        \bottomrule
	\end{tabular}
	}}
\end{table}
This experiment shows that the imprecision of the visualisation method used in ProtoPNet and ProtoTree leads to a false sense of model bias
(Table~\ref{tab:relevance}). In particular, when using the upsampling method, 35.4\% of ProtoTree prototypes and 51.9\% of the test image patches seem to be focusing on elements of the background rather than the bird. However, when using a more faithful saliency method such as PRP, we notice that only 0.5\% of the prototypes or test image patches are actually irrelevant. Note that this gap between methods is again more limited for ProtoPNet with CUB-c, where the upsampling method is less likely to ``miss'' the object entirely.
Finally, we also notice that the percentage of irrelevant prototypes and test image patches is significantly more important when using VGG19 as a backbone, compared to using Resnet50, which raises the question of the sensitivity of prototype-based architectures to the underlying backbone architecture.

\section{Discussion}\label{sec:discussion}
Case-based reasoning architectures for image classification are undoubtedly a stepping stone towards more interpretable computer vision models, but they are highly dependent on the choice of a backbone and suffer from shortcomings that may hinder their widespread usage.
Indeed, even when such models produce a correct decision for the right reasons, they may yet fail to explain this decision by incorrectly locating appropriate parts of the images. This issue is likely not restricted to ProtoPNet or ProtoTree, since ProtoPShare, ProtoPool, Deformable ProtoPNet and TesNet share a common code base inherited from ProtoPNet that includes the upsampling method for patch visualisation. On the contrary, more faithful saliency methods can help uncover biases~\cite{gautam2022this} or - in our experiments - disprove \textit{apparent} biases of the model.
Crucially, proving that the model is indeed focusing on the object does not imply that the decision is based on understandable information. Indeed, the assumption that proximity in the latent space entails perceptual similarity in the visual space may not always hold~\cite{hoffmann2021this,nauta2020this}.
Thus, we argue that a decision-making process based on distance in the latent space is not sufficient to guarantee interpretability, \ie that
case-based reasoning architectures using CNNs for feature extraction are currently not \textit{inherently} self-explainable.
Consequently, such models should not be compared according to their classification accuracy, but rather using metrics for evaluating various properties of explanations in a systematic manner (\eg \textit{sparsity}~\cite{gomez2022metrics}) and, in the case of prototype-based architectures, for quantifying visual similarity~\cite{wang2004image,nauta2020this}.



\paragraph{Acknowledgments} Experiments presented in this paper were carried out using the Grid'5000 testbed, supported by a scientific interest group hosted by Inria and including CNRS, RENATER and several Universities as well as other organizations (see https://www.grid5000.fr).\\
This work has been partially supported by MIAI@Grenoble Alpes, (ANR-19-P3IA-0003) and TAILOR, a project funded by EU Horizon 2020 research and innovation programme under GA No 952215.
\paragraph{Reproducibility} Our code is available at the following url: \href{https://github.com/romain-xu-darme/prototype_sanity_checks.git}{https://github.com/romain-xu-darme/prototype\_sanity\_checks.git}

\newpage


\end{document}


\title{Sanity checks for patch visualisation in prototype-based image classification\\
- Supplementary material -}

\author{Romain Xu-Darme$^{1,2}$, Georges Quénot$^2$, Zakaria Chihani$^1$, Marie-Christine Rousset$^2$\\ \\
$^1$ Université Paris-Saclay, CEA, List, F-91120, Palaiseau, France\\
$^2$ Univ. Grenoble Alpes, CNRS, Grenoble INP, LIG, F-38000 Grenoble, France \\
{\tt\small \{romain.xu-darme, zakaria.chihani(at)cea.fr\}}\\
{\tt\small \{georges.quenot, marie-christine.rousset(at)imag.fr\}}\\
}
\maketitle

\section{Training prototype-based classifiers}
In this work, we use the code provided by the authors of ProtoPNet~\cite{chen2019this} and ProtoTree~\cite{nauta2021neural} in order to train classifiers on the CUB-200-2011 dataset~\cite{welinder2010caltech} (CUB) and the StanfordCars dataset~\cite{krause20133d}. For ProtoPNet, we use the default training parameters provided by the authors and trained the models during 50 epochs. For ProtoTree, we also use the default parameters and trained the models during 100 epochs. Table~\ref{tab:accuracies} presents the final accuracy of the models used in our experiments.
\begin{table}[htbp]
\centering
\caption{\textbf{Accuracy of the self-explaining models} used in this work. CUB-c denotes the cropped CUB-200-2011 dataset.}\label{tab:accuracies}
\vspace{0.3cm}
	\begin{tabular}{@{\hskip 8pt}c@{\hskip 8pt}|@{\hskip 8pt}c@{\hskip 8pt}|@{\hskip 8pt}c@{\hskip 8pt}|@{\hskip 8pt}c}
		\toprule
		Model & Backbone & Dataset & Accuracy \\
		\hline
		\multirow{3}{*}{ProtoPNet} & VGG19    & CUB-c & 75.1\% \\
		\cline{2-4}
                                   & \multirow{2}{*}{ResNet50} & CUB-c & 72.5\% \\
                                                  	         & & CARS & 71.4\% \\
        \midrule
        \multirow{2}{*}{ProtoTree} & \multirow{2}{*}{ResNet50} & CUB & 83.1\% \\
                                                  	         & & CARS & 83.2\%\\
        \bottomrule
	\end{tabular}
\end{table}

\section{More prototype visualisation with Smoothgrads and PRP}
\begin{figure}
	\centering
    \includegraphics[width=0.8\linewidth]{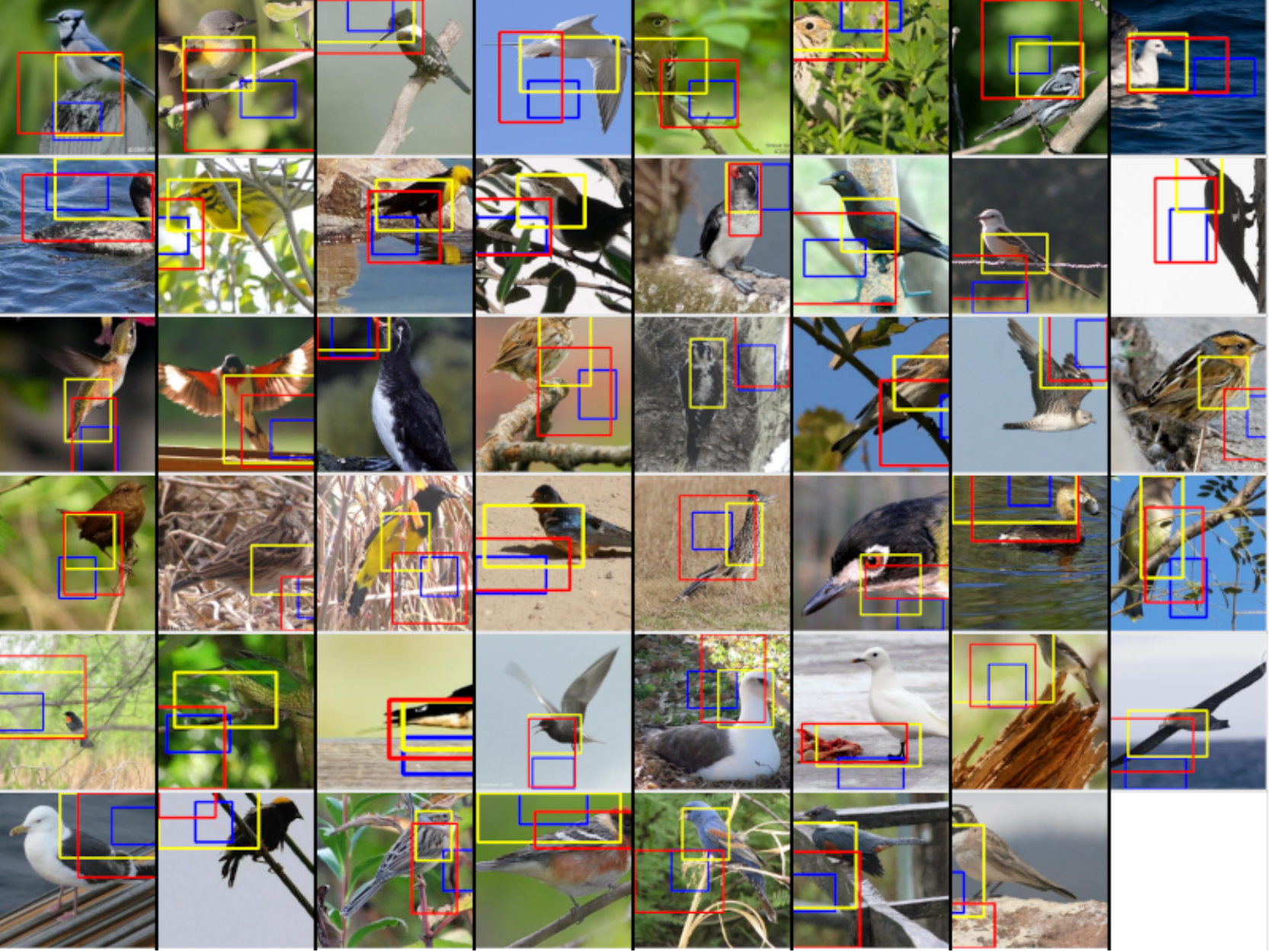}
    \caption{More examples of visualization of prototypes from a ProtoTree trained on CUB-200-2011 using upsampling with cubic interpolation (blue), Smoothgrads (red) or PRP (yellow).}\label{fig:proto_cmp_full}
\end{figure}

Fig.~\ref{fig:proto_cmp_full} illustrates how using more faithful visualisation methods, such as PRP or Smoothgrads, rather than upsampling can  improve the trust that the user can have in the model. In these examples, the upsampling strategy shows image patches focused on the background and gives a false sense of bias in the model, while PRP and Smoothgrads - which provide more faithful saliency maps - are focusing on elements of the bird.

\section{Area under the Deletion Curve}
\begin{figure*}
	\centering
	\includegraphics[width=0.8\linewidth]{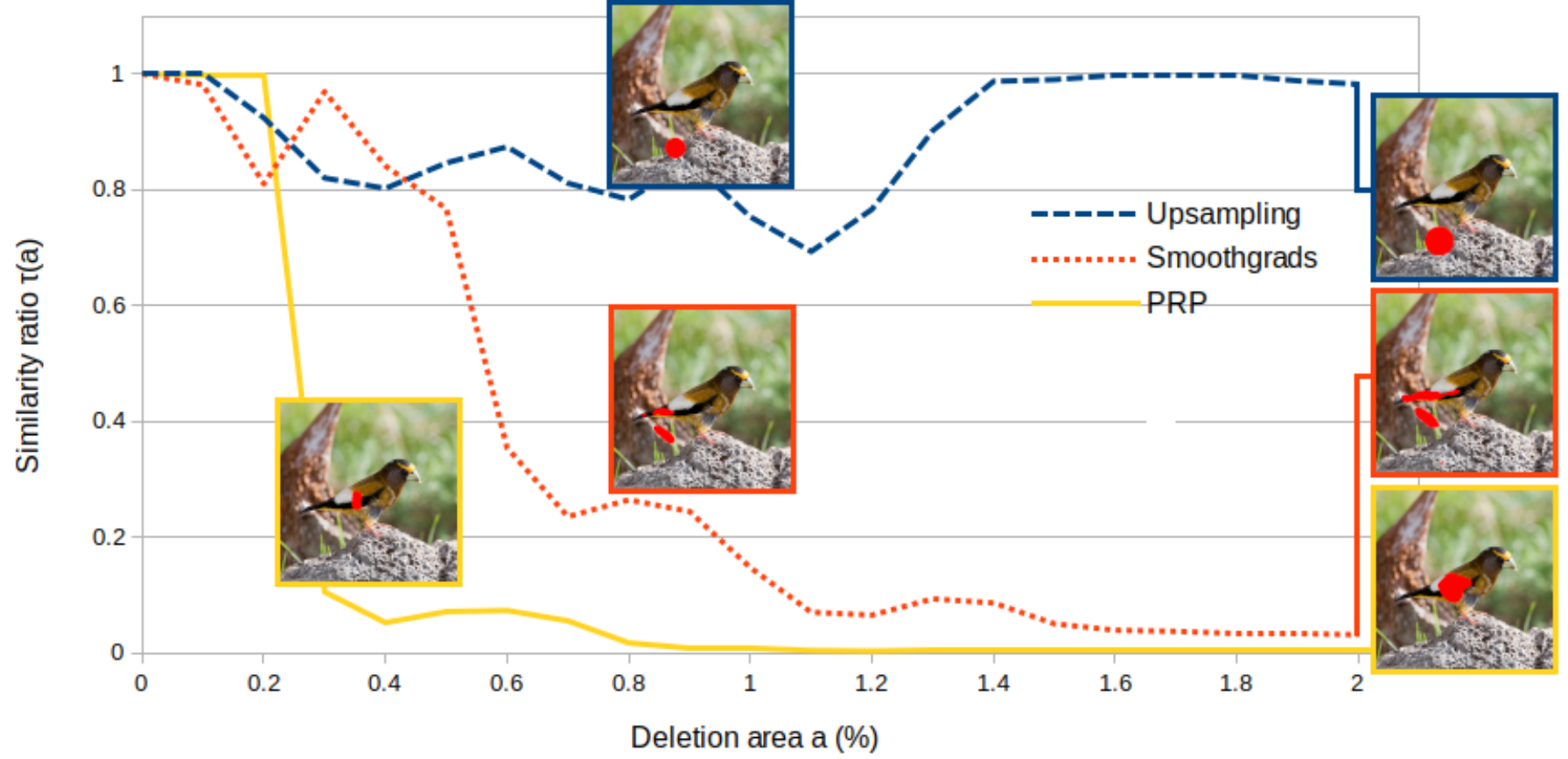}
	\caption{Evolution of the similarity ratio when incrementally removing the most important pixels according to the ProtoPNet/ProtoTree method (upsampling, in blue), Smoothgrads (red), and PRP (yellow). Best viewed in colour.}
	\label{fig:deletion_curve}
\end{figure*}
In this section, we illustrate the evolution of the similarity ratio when incrementally removing the most important pixels of the image according to the saliency maps proposed by the different visualisation methods under study. As shown in the example of Fig.~\ref{fig:deletion_curve}, removing pixels according to upsampling has little to no effect on the similarity score, suggesting that the explanation is incorrect. On the contrary, when removing only 1\% of the image according to Smoothgrads, the similarity score drops to roughly 15\% of its original value, suggesting that the explanation focuses on actual regions of interest for the model. The same result is achieved when removing only 0.3\% of the image according PRP, indicating an even more precise explanation. Moreover, reaching a similarity ratio lower than 10\% indicates that the explanation method has successfully identified the most relevant pixels of the image patch and gives an indication on the effective size of the image patch.

\section{Distribution of similarity ratio v. deletion area on ProtoTree visualisation}
\begin{figure*}
	\centering
	\includegraphics[width=0.8\linewidth]{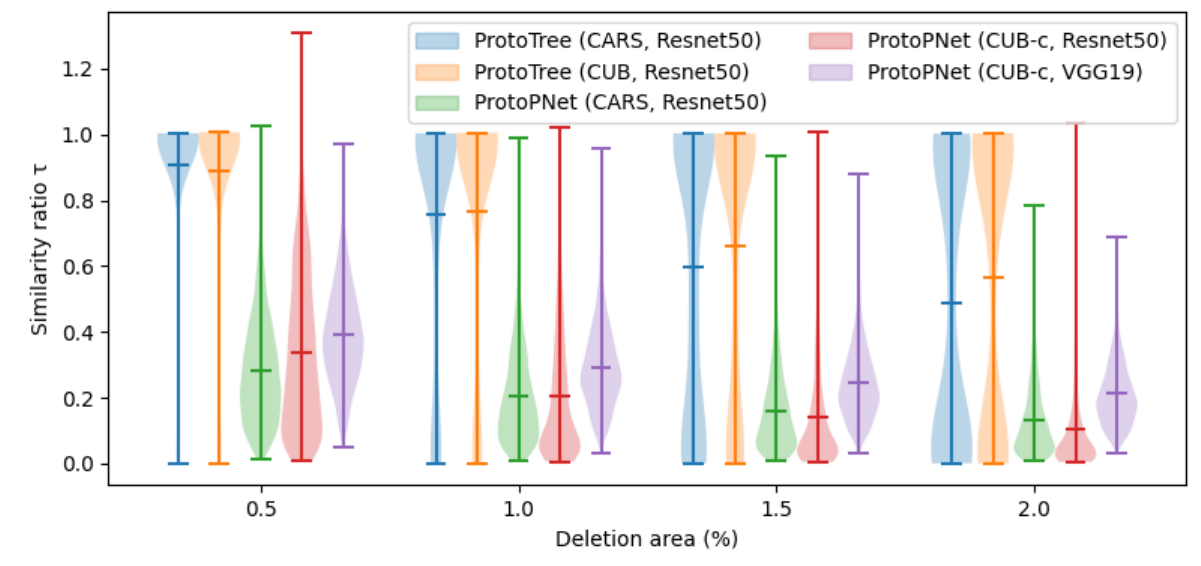}
	\caption{Distributions of similarity ratios v. percentage of deletion area when visualising prototypes using PRP.}
	\label{fig:prototype_violin_prp}
\end{figure*}

\begin{figure*}
    	\centering
    \begin{subfigure}[b]{0.8\linewidth}
    	\centering
    	\includegraphics[width=\linewidth]{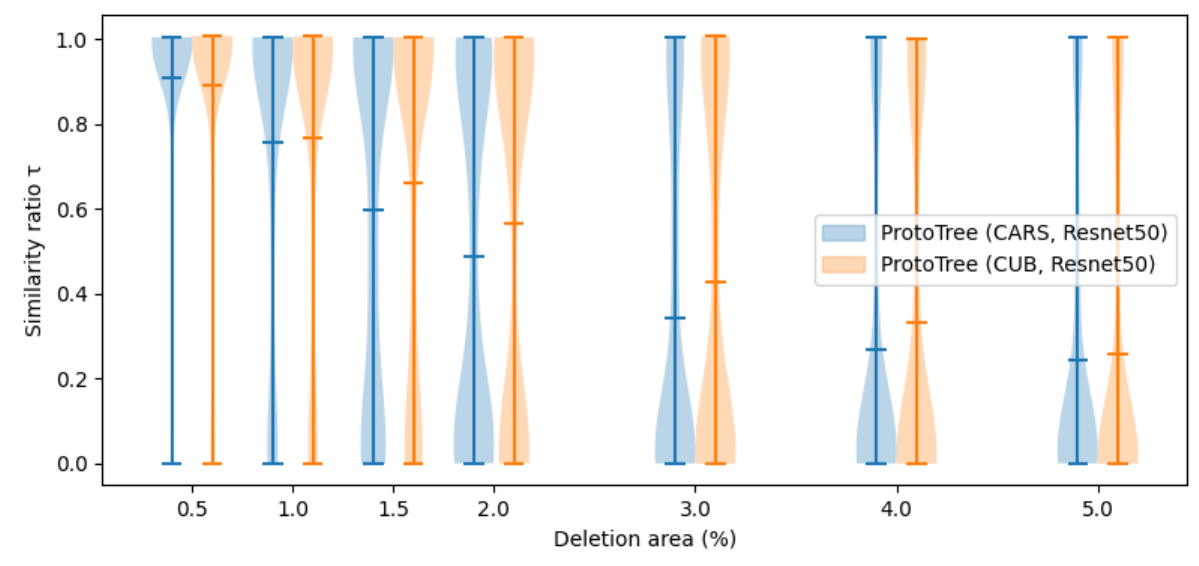}
    	\caption{Using PRP.}
    	\label{fig:prototree_prototype_violin_prp}
    \end{subfigure}
    \begin{subfigure}[b]{0.8\linewidth}
    	\centering
    	\includegraphics[width=\linewidth]{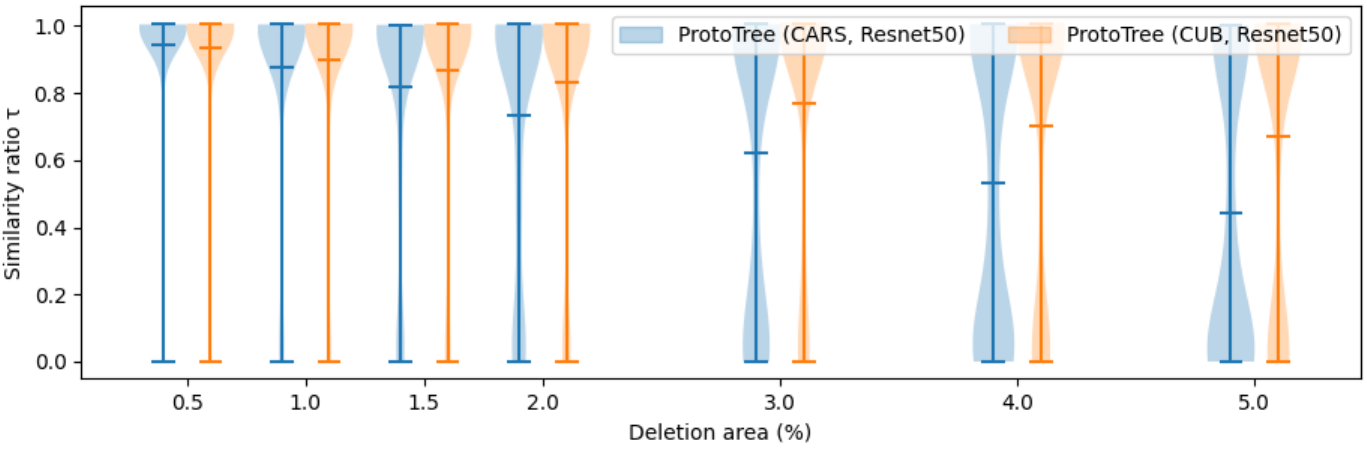}
    	\caption{Using Smoothgrads.}
    	\label{fig:prototree_prototype_violin_sm}
    \end{subfigure}
    \caption{Distributions of similarity ratios v. percentage of deletion area when visualising ProtoTree prototypes}
\end{figure*}

\begin{figure*}
	\centering
	\begin{subfigure}[b]{0.8\linewidth}
		\centering
		  \includegraphics[width=\linewidth]{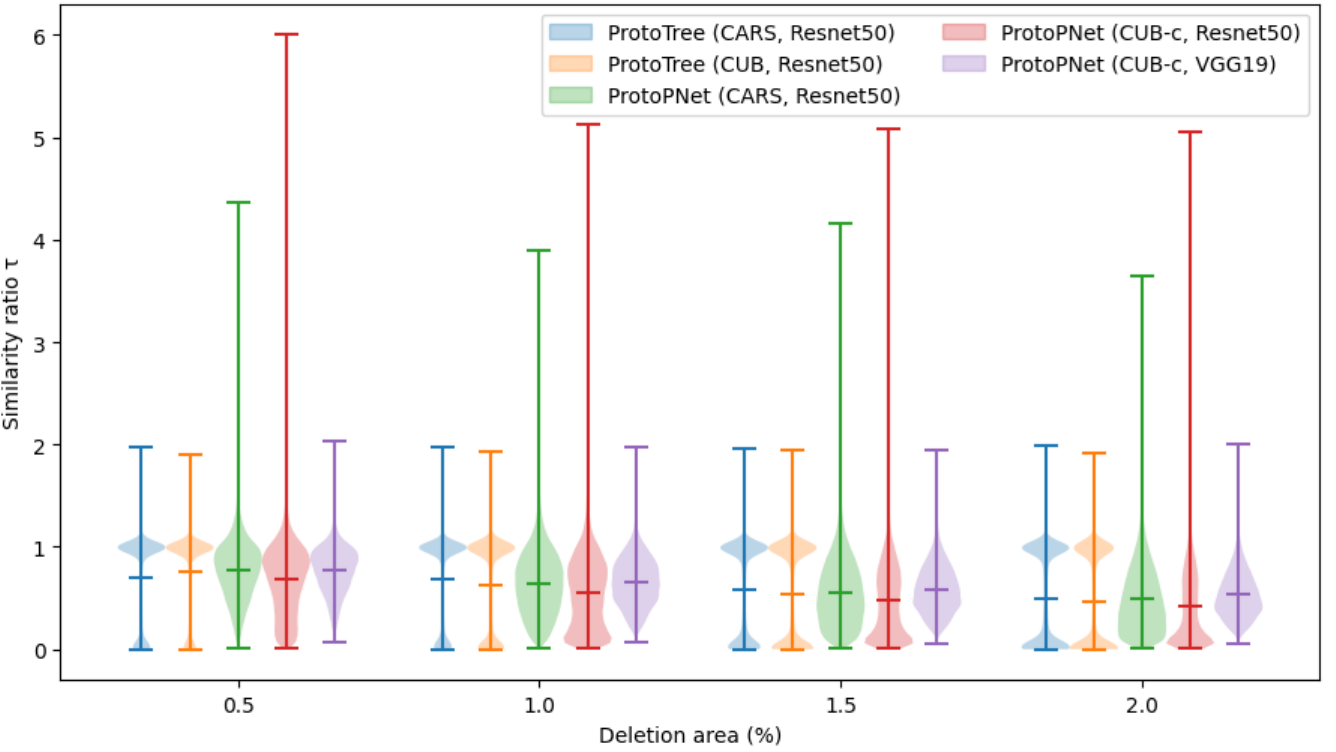}
			\caption{Using PRP}
	\end{subfigure}
	\hfill
	\begin{subfigure}[b]{0.8\linewidth}
		\centering
		  \includegraphics[width=\linewidth]{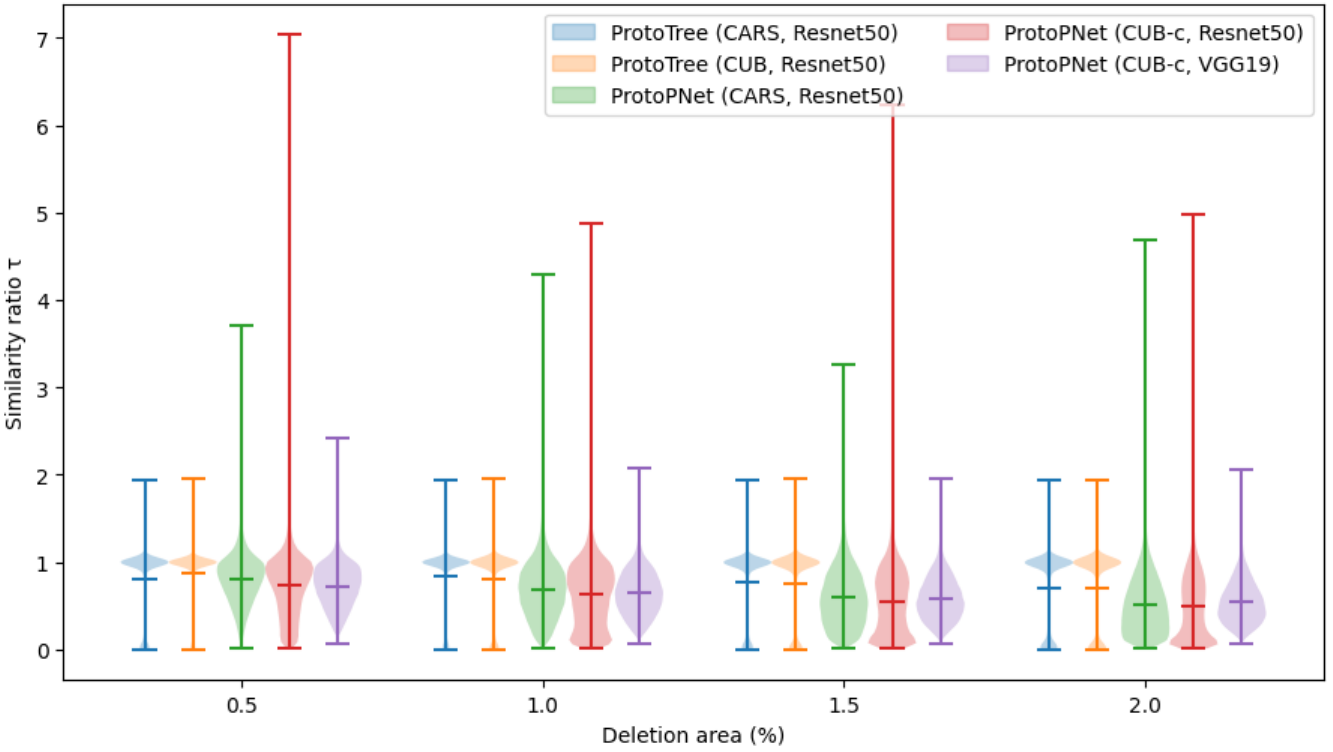}
			\caption{Using Smoothgrads}
	\end{subfigure}
	\caption{Distributions of similarity ratios v. percentage of deletion area when visualising test patches during inference.}\label{fig:general_inference_violin}
\end{figure*}
\begin{figure*}
\centering
	\begin{subfigure}[b]{0.8\linewidth}
		\centering
		  \includegraphics[width=\linewidth]{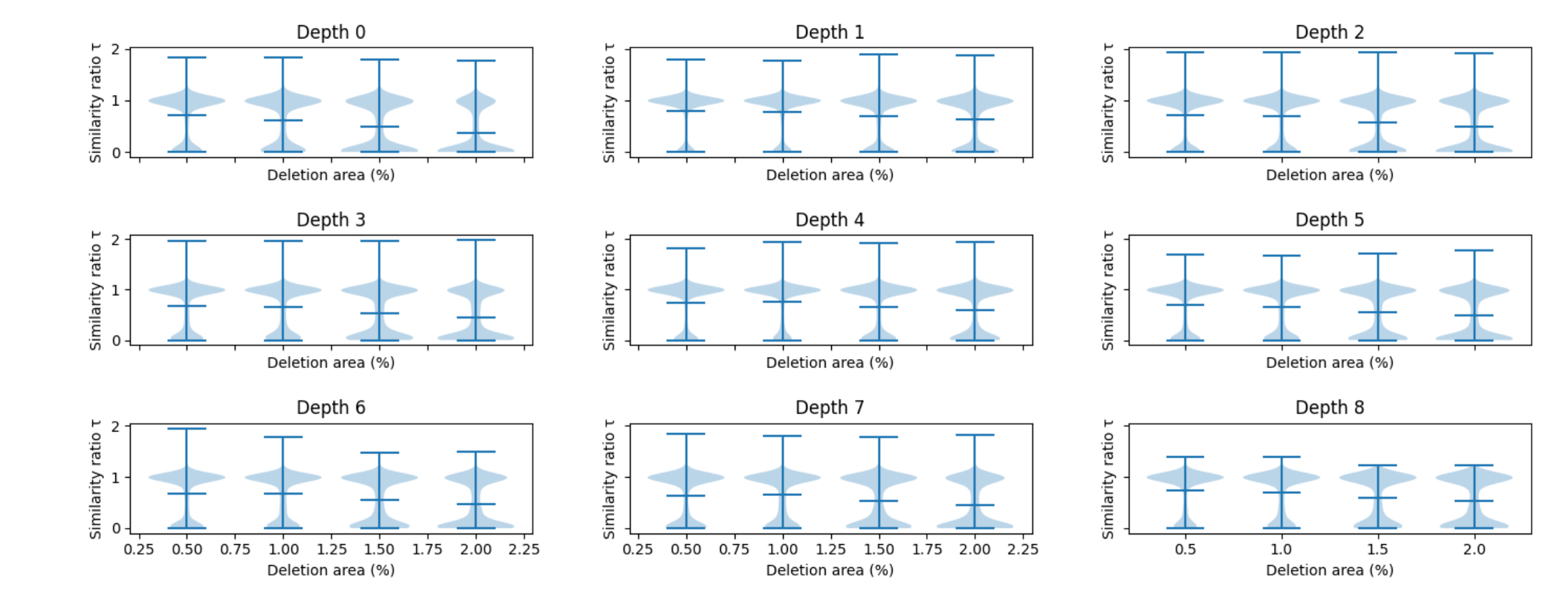}
			\caption{On CARS}
	\end{subfigure}
	\hfill
	\begin{subfigure}[b]{0.8\linewidth}
		\centering
		  \includegraphics[width=\linewidth]{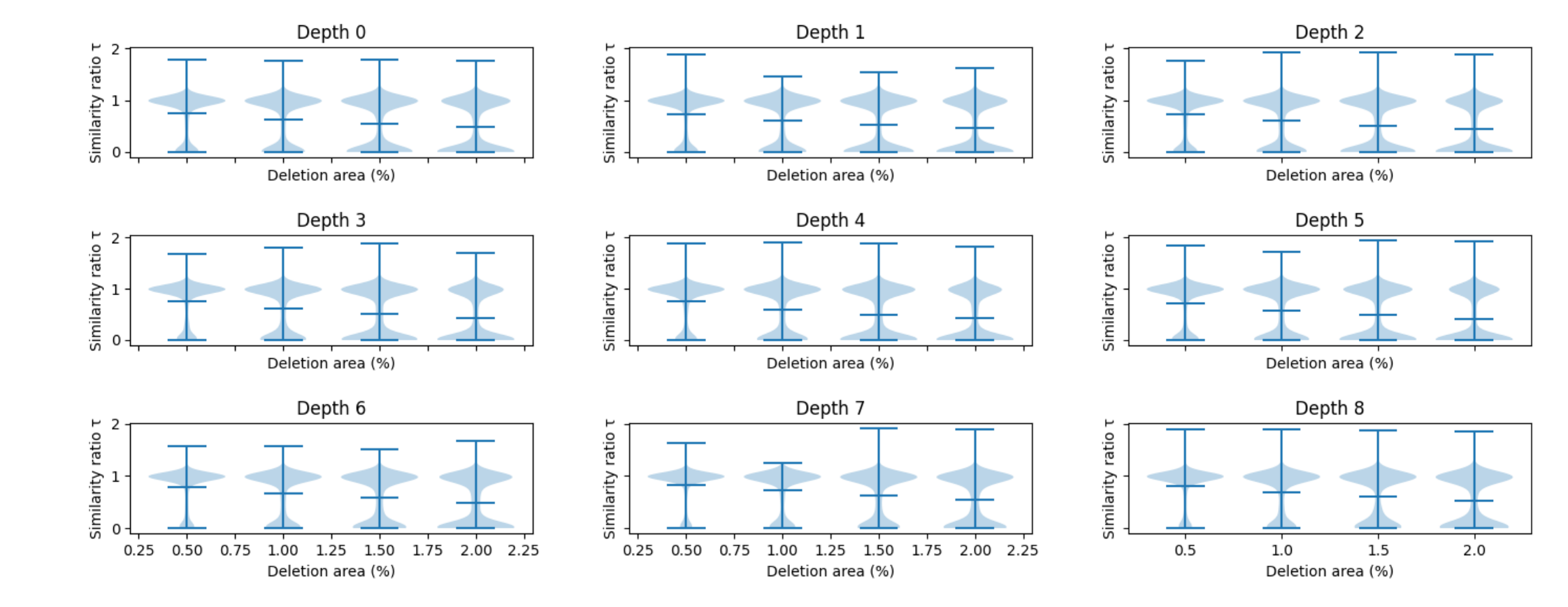}
			\caption{On CUB}
	\end{subfigure}
	\caption{Distributions of similarity ratios v. percentage of deletion area when visualising ProtoTree test patches using PRP during inference. Results are sorted by depth inside of the decision tree.}\label{fig:prototree_inference_violin_depth}
\end{figure*}
In addition to the results presented in the paper focusing on the average similarity ratio v. deletion area, in this section we study the \textit{distribution} of similarity ratios for a given deletion area (here 0.5\%, 1\%, 1.5\% and 2\%). As shown in Fig.~\ref{fig:prototype_violin_prp}, we notice a "sandglass" effect on the distribution of similarity ratios for ProtoTree prototypes: for low deletion areas ($\leq 1\%$), the similarity ratio for all prototypes is close to 1. Then, from 1.5\% up to 4-5\% (Fig.\ref{fig:prototree_prototype_violin_prp}), the distribution of similarity ratios slowly shifts towards 0. This suggests that the drop in similarity does occurs uniformly for all prototypes, but rather in a "continuous" manner, \ie that ProtoTree prototypes have a wider range of size for their corresponding effective receptive fields than ProtoPNet prototypes. Moreover, as shown in Fig.~\ref{fig:prototree_prototype_violin_sm}, the sandglass effect is also present when using Smoothgrads and when visualising image patches during inference (Fig.~\ref{fig:general_inference_violin}). Moreover, as shown in Fig.~\ref{fig:prototree_inference_violin_depth}, this effect is seemingly uncorrelated to the depth of the prototype inside of the decision tree. This suggests that ProtoTree does not necessarily focus on finer - and smaller - details in the last stages of the decision process.
\clearpage
{\small

}